\def\year{2022}\relax
\documentclass[letterpaper]{article} 
\usepackage{aaai22}  
\usepackage{times}  
\usepackage{helvet}  
\usepackage{courier}  
\usepackage[hyphens]{url}  
\usepackage{graphicx} 
\usepackage{natbib}  
\usepackage{caption} 
\usepackage{xcolor} 
\DeclareCaptionStyle{ruled}{labelfont=normalfont,labelsep=colon,strut=off} 
\usepackage{xcolor}

\usepackage{algorithm}
\usepackage{algorithmic}

\usepackage{newfloat}
\usepackage{listings}
\floatstyle{ruled}
\newfloat{listing}{tb}{lst}{}
\floatname{listing}{Listing}
\title{\textsc{RerrFact}: Reduced Evidence Retrieval Representations for Scientific Claim Verification}
\author{
    Ashish Rana\equalcontrib\textsuperscript{\rm 1}, Deepanshu Khanna\equalcontrib\textsuperscript{\rm 2}, Tirthankar Ghosal\textsuperscript{\rm 3}, \\
    Muskaan Singh\textsuperscript{\rm 3}, Harpreet Singh\textsuperscript{\rm 4}, Prashant Singh Rana\textsuperscript{\rm 4}
}
\affiliations{
    \textsuperscript{\rm 1}School of Business Informatics and Mathematics, University of Mannheim, Germany\\
    \textsuperscript{\rm 2}Department of Electronics \& Communication Engineering, Thapar Institute of Engineering \& Technology, India \\
    \textsuperscript{\rm 3}Institute of Formal and Applied Linguistics, Faculty of Mathematics and Physics, Charles University, Czech Republic \\
    \textsuperscript{\rm 4}Department of Computer Science \& Engineering, Thapar Institute of Engineering \& Technology, India \\
    asrana@mail.uni-mannheim.de\textsuperscript{\rm 1}, dkhanna\_be19@thapar.edu\textsuperscript{\rm 2}\\
    (ghosal, singh)@ufal.mff.cuni.cz\textsuperscript{\rm 3},
    (harpreet.s, prashant.singh)@thapar.edu\textsuperscript{\rm 4}
}

\usepackage{bibentry}

\begin{document}

\maketitle
\begin{abstract}
Exponential growth in digital information outlets and the race to publish has made scientific misinformation more prevalent than ever. However, the task to fact-verify a given scientific claim is not straightforward even for researchers. Scientific claim verification requires in-depth knowledge and great labor from domain experts to substantiate supporting and refuting evidence from credible scientific sources. The \textsc{SciFact} dataset and corresponding task provide a benchmarking leaderboard to the community to develop automatic scientific claim verification systems via extracting and assimilating relevant evidence rationales from source abstracts. In this work, we propose a modular approach that sequentially carries out binary classification for every prediction subtask as in the \textsc{SciFact} leaderboard. Our simple classifier-based approach uses reduced abstract representations to retrieve relevant abstracts. These are further used to train the relevant rationale-selection model. Finally, we carry out two-step stance predictions that first differentiate non-relevant rationales and then identify supporting or refuting rationales for a given claim. Experimentally, our system \textsc{RerrFact} with no fine-tuning, simple design, and a fraction of model parameters fairs competitively on the leaderboard against large-scale, modular, and joint modeling approaches. We make our codebase available at \textit{https://github.com/ashishrana160796/RerrFact}.
\end{abstract}
\vspace{- 2 ex}
\section{Introduction}
Misinformation is a modern day societal problem that has the potential to wreck havoc, especially with increasingly many people having an online footprint without adequate internet literacy. The problem grows intense when science gets associated with disinformation and provides a false sense of trustworthiness. Convincing statements derived from general public opinions like \textit{``Ginger consumption in food reduces the risk of getting severely infected with COVID-19"} can effectively manipulate the masses. It is hard to verify such misleading statements from extensive scientific literature with appropriate reasoning even by providing relevant evidence. Also, it is a cumbersome task for experts to search for refuting or supporting argument rationales considering the amount of misinformation available on a plethora of outlets. Therefore, automatic fact-verification tools are essential, especially for scientific knowledge where the given system must understand scientific knowledge, interpret numeric and statistical inferences.

\begin{figure*}
\centering
\includegraphics[width=\textwidth]{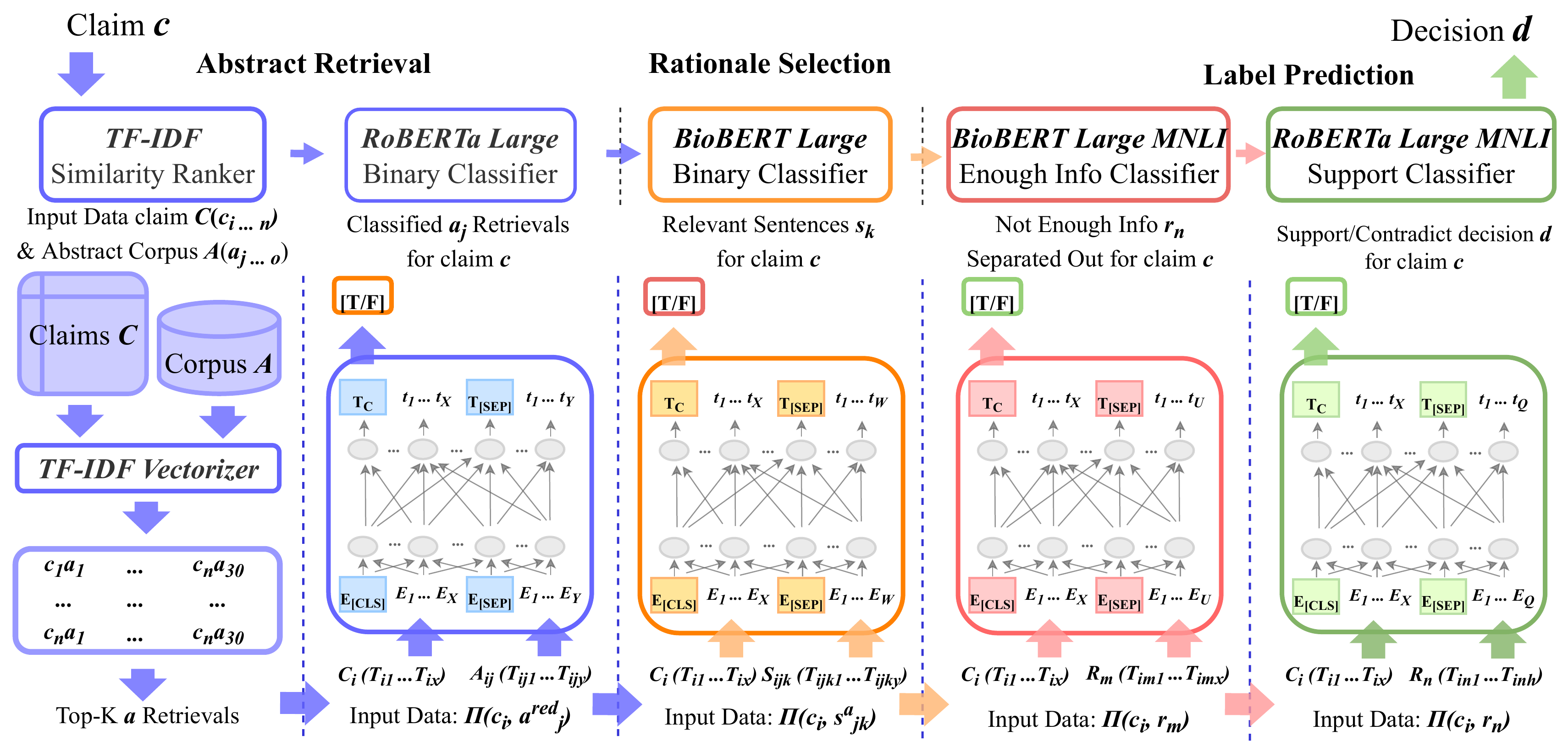}
\caption{Architectural illustration of the \textsc{RerrFact} scientific claim verification pipeline.}
\label{figure-one}
\end{figure*}

Previously, the veracity verification task has been extensively studied, and many datasets are available on various use-cases~\cite{deyoung-etal-2020-eraser, ferreira2016emergent, vlachos2014fact}. The most relevant amongst them is the FEVER shared task~\cite{thorne-etal-2018-fact}, which evaluates the veracity of human-generated claims from Wikipedia data. For the FEVER task, there are two paradigms: one that take a three-step modular approach and the other which is joint prediction approach for evidence retrieval \& stance prediction ~\cite{nie2019combining, chen-etal-2019-seeing}. Similarly, for the \textsc{SciFact} task these two paradigms have been used either with very large language models like \textsc{VerT5erini} for modular architecture~\cite{pradeep-etal-2021-scientific} or \textit{ARSJoint}, \textit{JointParagraph} for merged subtask architecture~\cite{DBLP:conf/aaai/LiBP21, zhang-etal-2021-abstract}. In contrast to these diametrically opposite paradigms, QMUL-SDS's ~\cite{zeng-zubiaga-2021-qmul} partial binding between the abstract retrieval and rational selection stages offers a promising direction, which is also the inspiration for our current work. Our experiments demonstrate that this partial interdependence successfully introduces a form of regularization, providing much-needed improvements over precision and recall for the evidence retrieval component in the concerned task. Therefore, we present a computationally and architecturally simple pipeline-driven design for it.

We use the same partial interdependence pipeline design with \textit{reduced evidence retrieval stage representations} for modeling our system \textsc{RerrFact}'s subtask modules. We also align our efforts to maximize performance from each subtask performing binary classification instead of opting for approaches like external data fine-tuning, utilizing extensive language models like T5, or using the joint learning architecture, etc. Here, we use the \textit{reduced abstract representations} after the initial TF-IDF retrieval for the relevant abstract extraction subtask. After that, we use these retrieved abstracts for training the rationale selection model that adds a loose coupling effect between the two evidence retrieval subtasks. Finally, for stance prediction, we first segregate out \{\textsc{NoInfo}\} rationale instances and then predict stance for \{\textsc{Supports}, \textsc{Refutes}\} rationales. \textsc{RerrFact} achieves the fourth rank in \textit{SciFact} leaderboard by using language models of different BERT-variants, choosing the best performing one for each subtask.  Our experimental results demonstrate the importance of this \textit{loose coupling phenomenon} as we only stand after computationally expensive approaches that require much larger language models and optimization for various thresholding parameters for each subtask.

\section{\textsc{SciFact} Dataset and Task Description}

The \textsc{SciFact} dataset consists of a corpus with 5,183 relevant abstracts for 1,409 scientific claims~\cite{wadden-etal-2020-fact}. These abstracts can either support or refute a claim with manually annotated rationales. Each claim has a unique single label, and no abstract has more than three rationales for a given claim. The natural claims derived from a paper and the papers cited in different paragraphs in it make the language modeling subtasks challenging especially due to added contextual scientific nuance.

For the \textsc{SciFact} task, one is given scientific claims $\mathcal{C}$ and a relevant abstract corpus $\mathcal{A}$~\cite{wadden-etal-2020-fact}. First, corresponding to a claim \textit{c} $\in$ $\mathcal{C}$, all unique abstracts \textit{a} $\in$ $\mathcal{A}$ are categorized as \textit{y(c,a)} $in$ \{\textsc{Supports}, \textsc{Refutes}, \textsc{NoInfo}\}. Second, the sentence selection task functionally retrieves the relevant rationales \{\textit{r\textsubscript{1}(c,a), …, r\textsubscript{m}(c,a)}\} $\in$ $\mathcal{R}$  for the given claim \textit{c} for each abstract \textit{a}. The performance of both these tasks is evaluated with \textit{precision, recall, and F1} metrics for abstract and sentence-level tasks. Third, for the veracity verification task which is formulated as a stance prediction problem, labels \{\textsc{Supports}, \textsc{Refutes}\} are considered as positive labels, and \{\textsc{NoInfo}\} is taken as the negative label.
\vspace{- 2 ex}

\section{Methodology}

We formulate each subtask for the \textsc{SciFact} task as a binary classification problem and create corresponding BERT representations for each sequence classifier. Figure~\ref{figure-one} depicts the summarized view of the proposed \textsc{RerrFact} system.

\subsection{Abstract Retrieval}

Here, we retrieve relevant abstracts from corpus \{\textit{a\textsubscript{1}, ..., a\textsubscript{j}}\} $\in$ $\mathcal{A}$ for claims \textit{c} $\in$ $\mathcal{C}$. First, we calculate the TF-IDF similarity of each claim c\textsubscript{i} with all abstracts a\textsubscript{o} in $\in$ $\mathcal{A}$ and restrict to top-\textit{K} (\textit{K} = 30) similar abstracts. Second, we create \textit{reduced abstract representations (a\textsuperscript{red} \textsubscript{j})} from these abstracts which is given by \textit{a\textsuperscript{red}\textsubscript{j}=\{title, s\textsubscript{1}, s\textsubscript{n/2}, s\textsubscript{n}\}}. These are empirically the most meaningful representations for RoBERTa large language model \cite{liu2020roberta}, which we use for binary classification with input sequence $<$ \textit{c\textsubscript{i},[SEP],a\textsuperscript{red} \textsubscript{j}}$>$ for obtaining all the relevant abstracts.

Additionally, we obtain the above-stated representation logic by permuting different combinations of abstract sentences. For all retrieval approaches, we append the title with different lengths of abstract. Keeping the language model architecture constant, for the baseline approach, we first feed the complete abstract \textit{a\textsuperscript{total}\textsubscript{j}} with the title into the model. But while appending the whole abstract due to the limitation of BERT models to take maximum 512 tokens as input on an average, our inputs get truncated, which possibly results in some information loss.

In the second approach, we divide our abstracts into different groups based on their sizes \textit{\{small($\leq$8*s\textsubscript{k}), medium($>$8*s\textsubscript{k} \& $\leq$14*s\textsubscript{k}),  large($>$14*s\textsubscript{k} \& $\leq$24*s\textsubscript{k}),  extra-large($>$24*s\textsubscript{k} \& $\leq$L\textsubscript{max}*s\textsubscript{k})\}}, and for each group of abstracts formed, we consider the top five relative index positions of the most frequently occurring sentences for each group and sequentially append those five sentences after the title \textit{(a\textsuperscript{diff-5}\textsubscript{j})} as our new input sequence to fine-tune our language model. Also, we follow the same methodology but limit our sentences to only top-three sentences appended after the title \textit{(a\textsuperscript{diff-3}\textsubscript{j})} for observing performance and computational trade-off variations on smaller representations.

\begin{table}[]
\scriptsize
    \centering
    \begin{tabular}{c|c}
         Abstract Classification Approach&F1-score\\
        \hline \\
        Total Abstract \textit{(a\textsuperscript{total}\textsubscript{j})}&72.25\\
        Diff-Size Abstracts, Five Sentences \textit{(a\textsuperscript{diff-5}\textsubscript{j})}&74.41\\
        Diff-Size Abstracts, Three Sentences \textit{(a\textsuperscript{diff-3}\textsubscript{j})}&68.63 \\
        \textsc{RerrFact}'s Reduced Abstract \textit{(a\textsuperscript{red}\textsubscript{j})}&\textbf{79.67}\\
    \end{tabular}
    \caption{F1-score performances on \textit{dev} set for different comparative abstract representations.}
    \label{Table a}
\end{table}

The results from Table \ref{Table a} demonstrate our final reduced retrieval representations outperforming other representations with its best F1-score. Our manual analysis into workings of these representations shows that the \textit{a\textsuperscript{red}\textsubscript{j}=\{title, s\textsubscript{1}, s\textsubscript{n/2}, s\textsubscript{n}\}} method captures qualitatively best portions of the introduction, methodology \& conclusion on an average. More importantly, unlike other approaches, it avoids the abstract's numeric \& additional bulk information components, keeping the representations compact \& precise.

\subsection{Rationale Selection}

In this subtask, relevant evidence rationales $\mathcal{\widehat{R}}$\textit{(c,a)} = \{\textit{r\textsubscript{1}(c,a), ..., r\textsubscript{m}(c,a)}\} are retrieved, where each \textit{r\textsubscript{1}(c,a)} comprises of \{\textit{s\textsubscript{1}(c,a), ..., s\textsubscript{k}(c,a)}\}. We use all sentences from each retrieved abstract from the previous stage to fine-tune our pre-trained BioBERT large language model \cite{lee2020biobert} with input sequence $<$\textit{c\textsubscript{i},[SEP],s\textsuperscript{a} \textsubscript{k}}$>$ and binary output \textit{[T/F]}. Binding the abstract retrieval module to the rationale selection module while model training helps in improving \textit{co-reference} identification performance and gives special attention only to claim relevant data.

Also, we further analyze different training mechanisms for the sentence selection subtask. First, we train our baselines only by using oracle retrieved abstract. Further, as a new variation, we add negative label sentences for claims with no supporting/refuting evidence but only respective \textit{cited\_doc\_id} in the abstract corpus. Second, we decide to add more negative samples by adding top-three falsely retrieved abstracts from initial TF-IDF similarity retrieval. Finally, we try our loose-coupling approach by binding training to classified abstracts only. The results from Table \ref{Table c} demonstrate the importance of the binding mechanism \& emphasize that adding negative samples does not necessarily improve results.

\begin{table}[]
\scriptsize
    \centering
    \begin{tabular}{c|c}
        Sentence Selection Approach & F1-score\\
        \hline \\
        BioBERT large Oracle Retrieval&67.63\\
        Oracle Retrieval + No Evidence \& Cited&65.47\\
        Oracle + No Evidence Cited + (-3)*TF-IDF &62.23\\
        \textsc{RerrFact}'s Loose Coupling&\textbf{69.57}\\
    \end{tabular}
    \caption{F1-score performances on \textit{dev} set for different comparative sentence selection approaches.}
    \label{Table b}
\end{table}

\subsection{Stance Prediction}

In this subtask, we use the predicted rationales $\mathcal{\widehat{R}}$\textit{(c,a)} = \{\textit{r\textsubscript{1}(c,a), …, r\textsubscript{m}(c,a)}\} from the evidence retrieval stage to predict the veracity \^{y}\textit{(c,a)} of the scientific claims \textit{c} $\in$ $\mathcal{C}$. We formulate this subtask as a two-stage binary classifier problem where the first classifier separates the rationales with \^{y}\textit{(c,a)}=\{\textsc{NoInfo}\} with input sequence $<$\textit{c\textsubscript{i},[SEP],r\textsubscript{m}}$>$ and the second classifier predicts the stance \^{y}\textit{(c,a)}=\{\textsc{Supports}, \textsc{Refutes}\} with input representation $<$\textit{c\textsubscript{i},[SEP],r\textsubscript{n}}$>$. We choose the pre-trained BioBERT-MNLI language model for \textit{Enough Information} detection and pre-trained RoBERTA-Large-MNLI for predicting \textit{Claim Veracity}.

Further, we explore three-way classification by training the individual models of the \textsc{RerrFact} veracity verification two-step module. We train our multiclass language model classifiers namely, BioBERT-MNLI \& RoBERTa-Large-MNLI  for directly predicting the  \{\textsc{Supports}, \textsc{Refutes}, \textsc{NoInfo}\} labels. The results in Table \ref{Table b} demonstrates the advantage of using the two-step binary classification process in \textsc{RerrFact} for the \textsc{SciFact} task. We attribute this performance increase to better prediction of \textsc{Refutes} class, as multiclass classification models performed poorly for predicting this class due to its scarcity in the dataset. Hence, \textsc{RerrFact}’s two-step classification approach avoids false positive predictions of \textsc{NoInfo} class against the \textsc{Refutes} class and improves on the claim refuting rationale prediction.

\begin{table}[]
\scriptsize
    \centering
    \begin{tabular}{c|c}
        Stance Prediction Approach & F1-score\\
        \hline \\
         BioBERT-MNLI \textit{(Multiclass)}&74.09\\
        RoBERTA-Large-MNLI \textit{(Multiclass)}&76.58\\
        \textsc{RerrFact}'s \textsc{NoInfo} \textit{(Binary)} &\textbf{87.14}\\
        \textsc{RerrFact}'s \textsc{Supports/Refutes} \textit{(Binary)}&\textbf{82.67}\\
        \textsc{RerrFact} Classifier \textit{(Two-Step Binary)}&\textbf{85.23}\\
    \end{tabular}
    \caption{F1-score performances on \textit{dev} set for different comparative stance prediction approaches.}
    \label{Table c}
\end{table}

\section{Experiment and Results}

\begin{table*}[]
\scriptsize
    \centering
    \begin{tabular}{c|ccc|ccc|ccc|ccc}
    \hline
         & \multicolumn{6}{c}{Sentence-level} &  \multicolumn{6}{c}{Abstract-level} \\
         & \multicolumn{3}{c}{Selection-only} & \multicolumn{3}{c}{Selection+Label} & \multicolumn{3}{c}{Label-Only} & \multicolumn{3}{c}{Label+Rationale} \\
         Models & P & R & F1 & P & R & F1 & P & R & F1 & P & R & F1 \\
         \hline \\
        \textsc{RerrFact} & \textbf{93.65} & \textbf{64.48} & \textbf{76.37} & \textbf{78.17} & \textbf{53.83} & \textbf{63.76} & 79.17 & 54.55 & 64.59 & \textbf{78.47} & 54.07 & \textbf{64.02} \\
        ARSJoint & 76.2 & 58.5 & 66.2 & 66.5 & 51.1 & 57.8 & 75.3 & 59.8 & \textbf{66.7} & 70.5 & 56.0 & 62.4 \\
        \textsc{VerT5erini} & 64.81 & 57.37 & 60.87 & 60.8 & 53.83 & 57.1 & 65.07 & \textbf{65.07} & 65.07 & 61.72 & \textbf{61.72} & 61.72 \\
        ParagraphJoint & 74.2 & 57.4 & 64.7 & 63.3 & 48.9 & 55.2 & 71.4 & 59.8 & 65.1 & 65.7 & 55.0 & 59.9 \\
         QMUL-SDS & 80.75 & 58.47 & 67.83 & 72.08 & 52.19 & 60.54 & \textbf{79.71} & 52.63 & 63.40 & 76.81 & 50.72 & 61.10 \\
         \textsc{VeriSci} & 54.3 & 43.4 & 48.3 & 48.5 & 38.8 & 43.1 & 56.4 & 48.3 & 52.1 & 54.2 & 46.4 & 50.0 \\
         \hline
    \end{tabular}
    \caption{\textsc{RerrFact}'s performance on \textsc{SciFact} tasks  on \textit{dev} set.}
    \label{Table 1}
\end{table*}

\begin{table*}[]
\scriptsize
    \centering
    \begin{tabular}{c|ccc|ccc|ccc|ccc}
    \hline
         & \multicolumn{6}{c}{Sentence-level} &  \multicolumn{6}{c}{Abstract-level} \\
         & \multicolumn{3}{c}{Selection-only} & \multicolumn{3}{c}{Selection+Label} & \multicolumn{3}{c}{Label-Only} & \multicolumn{3}{c}{Label+Rationale} \\
         Models & P & R & F1 & P & R & F1 & P & R & F1 & P & R & F1 \\
         \hline \\
         \textsc{VerT5erini} & 63.05 & 69.19 & 65.98 & 60.59 & \textbf{66.49} & \textbf{63.40} & 64.03 & \textbf{72.97} & 68.21 & 62.85 & \textbf{71.62} & 66.95 \\
         ARSJoint & 79.53 & \textbf{72.43} & \textbf{75.81} & 66.17 & 60.27 & 63.08 & 72.22 & 64.41 & 68.10 & 69.70 & 62.16 & 65.71 \\
         \textsc{RerrFact} & 80.07 & 58.65 & 67.71 & \textbf{73.43} & 53.78 & 62.09 & \textbf{82.89} & 56.76 & 67.38 & \textbf{81.58} & 55.86 & 66.31 \\
         ParagraphJoint & 79.86 & 63.24 & 70.59 & 68.94 & 54.59 & 60.94 & 75.81 & 63.51 & \textbf{69.12} & 73.66 & 61.71 & \textbf{67.16} \\
         QMUL-SDS & \textbf{81.58} & 58.65 & 68.24 & 66.17 & 47.57 & 55.35 & 74.32 & 49.55 & 59.46 & 72.97 & 48.65 & 58.38 \\
         \textsc{VeriSci} & 44.99 & 47.30 & 46.11 & 38.56 & 40.54 & 39.53 & 47.51 & 47.30 & 47.40 & 46.61 & 46.40 & 46.50 \\
         \hline
    \end{tabular}
    \caption{\textsc{RerrFact}'s performance on \textsc{SciFact} tasks  on \textit{test} set.}
    \label{Table 2}
\end{table*}

In our experiments, we analyze the performance of various language models in a standalone manner for each subtask and attempt multiple permutation settings for our system \textsc{RerrFact} as shown in Tables \ref{Table a}, \ref{Table b} and \ref{Table c}. Table \ref{Table 1} and Table \ref{Table 2} report the performance of our best language models in \textsc{RerrFact} for each subtask in \textit{SciFact} against the top leaderboard systems on both \textit{dev} and \textit{test} sets. For evaluation and reporting performance on the \textit{dev} set, all language models for each subtask are trained only on the \textit{train} set. Table \ref{Table 1} shows the evaluation results against the \textit{dev} set having 300 claims. And for evaluation against the \textit{test} set predictions, we train our models on the \textit{train} set additionally combined with 75\% of the \textit{dev} set and validate our model results over the remaining 25\% of the \textit{dev} set. Table \ref{Table 2} reports the \textsc{RerrFact} system's capabilities in terms of $F_1$ scores against 300 claims of the \textit{test} set.

\begin{table*}[]
\scriptsize
    \centering
    \begin{tabular}{ll}
        \hline \\
        \textbf{Scientific Claim } \textit{(Reasoning Type, Frequency \%)} & \textbf{Wrongly Labeled Evidence} \textit{(Stance Gold Label)} \\
        \hline \\
        
         \textit{1/2000 in UK have abnormal PrP positivity.}
           &  \textit{...indicating an overall prevalence of 493 per million population} \\
           \textbf{\textit{(Numeric, 27.7\%)}} & \textit{(95\% confidence interval 282 to 801 per million)...} \textcolor{green}{\textbf{\textit{\textsc{\{Support\}}}}}  \\
        \hline \\

         \textit{Hypothalamic glutamate neurotransmission is}
           &  \textit{...secondary to impaired fasting-induced increases in the glucose-} \\
           \textit{crucial to energy balance.} \textbf{\textit{(Directionality, 37.9\%)}} & \textit{raising pancreatic hormone glucagon and...} \textcolor{green}{\textbf{\textit{\textsc{\{Support\}}}}}  \\
        \hline \\

         \textit{Breast cancer development is determined}
           &  \textit{...women who developed breast cancer... established environmental} \\
           \textit{exclusively by genetic factors.} \textbf{\textit{(Causal Effect, 34.4\%)}} & \textit{risk factors...alcohol consumption).} \textcolor{red}{\textbf{\textit{\textsc{\{Contradict\}}}}}  \\
        \hline \\
        
    \end{tabular}
    \vspace{- 2 ex}
    \caption{Reasoning categories where \textsc{RerrFact} fails to predict correct labels.}
    \label{Table 3}
    \vspace{- 2 ex}
\end{table*}

In the \textit{abstract retrieval} subtask, we empirically observe that the \textit{reduced abstract representations} substantially increase our retrieval performance, leading to a performance boost across all metrics in \textsc{SciFact}. This model is trained with batch size one for ten epochs. We achieve an \textit{F1-score} of 79.67\% against the \textit{dev} set, which is higher than reported QMUL-SDS's \textit{F1-score} of 74.15\% but lower than \textsc{VerT5erini}'s 89.95\% \textit{F1-score}. Second, for the \textit{rationale selection} subtask, the BioBERT-large language model attains a higher recall score in the \textsc{SciFact} metrics because of the loose binding between the two subtasks for evidence retrieval as part of \textsc{RerrFact}’s system design. Though our \textit{F1-score} performance for sentence selection was 69.57\% which is again less than \textsc{VerT5erini}'s \textit{F1-score} of 76.14\%, our performance on \textit{dev} set supersedes all the systems, including the T5 language models of \textsc{VerT5erini}. Based on our analysis of predictions from abstract and sentence selection subtasks, this performance boost largely attributes to the regularization effect created by loosely binding the two evidence retrieval stages leading to highly accurate sentence predictions for the retrieved abstracts.

For the final \textit{stance prediction} subtask, we train both our models in the two-step approach for 30 epochs with batch size 1. First, the \{\textsc{NoInfo}\} detector language model that eliminates evidence based on their unrelatedness to the scientific claim, achieves \textit{F1-score} of 87.14\%. The second stance predictor model for evidence that either supports or refutes the claim, achieves an \textit{F1-score} of 82.67\%. These two-step binary classifiers for neutral and support/refute evidence classification helps in achieving significant relative performance improvements on the \textit{dev} set, as shown in Table \ref{Table 1}'s label prediction metrics. Also, from Table \ref{Table 2}, we observe that \textsc{Rerrfact}’s performance takes a relatively large dip in terms of prediction capabilities because of the relatively lower abilities to detect true negatives for each subtask and wrong predictions on scientifically exhaustive rationales.
\section{Analysis}
Our manual analysis shows that \textsc{RerrFact}’s increase in performance can be attributed to its ability to process \textit{scientific background knowledge} and \textit{co-references} more accurately. First, the reduced abstract representations help in qualitatively improving the \textit{co-references} inference capabilities. Second, the dynamic biological pre-trained embeddings in classifier models help in increasing the \textit{scientific background knowledge}. Additionally, by coupling the sentence selection module’s training with retrieved abstract sentences as input, we add a form of regularization that increases generalization for rationale extraction subtask while keeping our sentence selection model compact. But, our system still fails to comprehend concepts like \textit{quantitative directionality},  \textit{numerical reasoning}, and \textit{causal effects}. This we further demonstrate by examples in Table \ref{Table 3} alongside their corresponding error-occurring frequency in \textit{dev} set over 29 misclassified \textit{claim-rationale} pairs.
\section{Conclusion}
In this work, our proposed system \textsc{RerrFact} demonstrates that reduced evidence retrieval representations and loosely binding the evidence retrieval stages for flexible regularization lead to better and concise retrieved rationale sentences. Additionally, combined with \textsc{RerrFact}’s two-step stance prediction approach, it outperforms all the other veracity verification systems on the \textsc{SciFact} \textit{dev} set. Also, for \textsc{RerrFact}, the performance especially takes a relatively high dip on the \textit{test} set, which can be attributed to a high false-positive rate on the \textit{test} set \& also that \textsc{SciFact} metric penalizations requiring more regularized predictions for each subtask. Our proposed system \textsc{RerrFact} ranks 4\textsuperscript{th} on the \textsc{SciFact} leaderboard, with 62.09\% F1-score for the Sentence+Label prediction module, while the top-performing system has an F1-score of 67.21\%. As future work, we would systematically improve upon these limitations and further explore novel premise assimilation architectures to create qualitatively improved veracity verification systems.
\bibliography{main}
\end{document}